# Evaluating AI-Driven Automated Map Digitization in QGIS

*Diana Febrita, Purdue University*

**Abstract**

Map digitization is an important process that converts maps into digital formats that can be used for further analysis. This process typically requires a deep human involvement because of the need for interpretation and decision-making when translating complex features. With the advancement of artificial intelligence, there is an alternative to conducting map digitization with the help of machine learning techniques. Deepness, or Deep Neural Remote Sensing, is an advanced AI-driven tool designed and integrated as a plugin in the QGIS application. This research focuses on assessing the effectiveness of Deepness in automated digitization. This study analyses AI-generated digitization results from Google Earth imagery and compares them with digitized outputs from OpenStreetMap (OSM) to evaluate performance.

## Introduction

Geospatial Artificial Intelligence or GeoAI is an integration of artificial intelligence (AI) with spatial data, science, and geospatial technology to improve understanding and solve spatial problems (Gao, et.al., 2023). This technology enables the automation of extracting information from satellite imagery, LiDAR, and unstructured texts to accelerate data processing and improve accuracy in the analysis. It opens new opportunities and innovation in many sectors. For example, it can help to manage infrastructure and optimize spatial planning. This technology can help predict disasters and speed up the emergency response as well.

GeoAI tools have been widely developed and used for many purposes in recent times. However, some tools can be challenging for non-expert machine learning users. Because of these reasons, this research will discuss an AI-driven tool that is designed for automated digitization. Deepness or Deep Neutral Remote Sensing is an open-source plugin in the QGIS software that can help detect and digitize objects automatically. This plugin enables data processing using provided deep learning models (Aszkowski, et.al., 2023).

Deepness is designed to facilitate the integration of deep learning models in geospatial analysis. This plugin allows users to apply various computer vision techniques, such as image classification, object detection, and others. Using custom ONNX Neural Network models, users can perform analysis faster and easier. However, the pre-trained deep learning models have not been universally trained. There may be variations in precision at different locations (QGIS, n.d). This research evaluates the performance of Deepness, a QGIS plugin for automated digitization by comparing its results with manually digitized data from human contributors in a location in the United States.

**Methods**

This study compares AI-based digitization using Deepness, a QGIS plugin, with human-digitized data extracted from OpenStreetMap (OSM) as the ground truth. Google Earth imagery serves as the primary data source for AI extraction, while OSM provides manually contributed geographic features. The AI-generated vector data and the human-digitized dataset are preprocessed and compared to evaluate the AI's performance in digitization.

The study employs QGIS, R, and Google Earth for data extraction, preprocessing, and analysis. Deepness in QGIS is used to identify and digitize geographic features automatically. R is utilized to extract and process data from OpenStreetMap (OSM) for comparison. The findings provide valuable insights into the feasibility of AI-driven digitization as an alternative to traditional digitization methods.

The study area is a residential neighborhood along Cumberland Avenue in West Lafayette, Indiana, United States. A total of 320 buildings from OpenStreetMap are compared with those digitized using an AI-based approach. Additionally, different parameter configurations are tested to assess their impact on the quality of AI-generated digitization.

To evaluate the AI tool's reliability, the F1-score was used, comparing AI-generated results to OSM ground truth. True positives (TP) signified buildings correctly identified by the AI in agreement with OSM, false positives (FP) were AI-detected buildings absent in OSM, and false negatives (FN) were OSM buildings missed by the AI. According to Powers (2011), true and false positives (TP/FP) indicate the number of predicted positives that were correct and incorrect, respectively, and similarly for true and false negatives (TN/FN).

Intersection over Union (IoU) is used to determine the true positives, false positives, and false negatives. If the overlap area is greater than 50%, the building is classified as a true positive. If the overlap area is less than 50%, the building is classified as a false positive. If the overlap area is zero, meaning the AI fails to detect the ground-truth building, the building is classified as a false negative. The formula for the F1 score is:

$$F_1 = 2\frac{Precision \times Recall}{Precision + Recall}$$

Where:

- Intersection Over Union (IoU)= Area of Intersection / Area of Union
- Precision = TP / (TP+FP)
- Recall = TP / (TP+FN)

Definitions:

- True Positive (TP): Detected buildings that match the ground truth with the IoU of more than 50%.
- False Positive (FP): Detected buildings that do not exist in the ground truth, or the IoU is less than 50%.
- False Negative (FN): Buildings in the ground truth that were missed in detection with IoU = 0.

## Results and Discussion

### *a. Deepness Digitization Settings*

The Deepness plugin enables users to automate the digitization process efficiently. However, before AI processing, users must configure several key parameters to achieve the best results. These include resolution, the percentage of tile overlap, and the option to remove or retain small segments. Properly adjusting these settings is important, as these parameters will determine the quality of the digitization results obtained.

This study used Deepness to digitize Google Earth imagery for a portion of the Cumberland Ave residential area in West Lafayette. Two resolution settings were tested: 100 cm/px and 300 cm/px, both with a consistent tile overlap of 12%. The results showed that the 100 cm/px resolution produced a significantly higher number of digitized buildings (2,909 buildings, as shown in Figure 1) compared to the 300 cm/px resolution (776 buildings, as shown in Figure 2). The 100 cm/px imagery also contained a greater number of small segments.

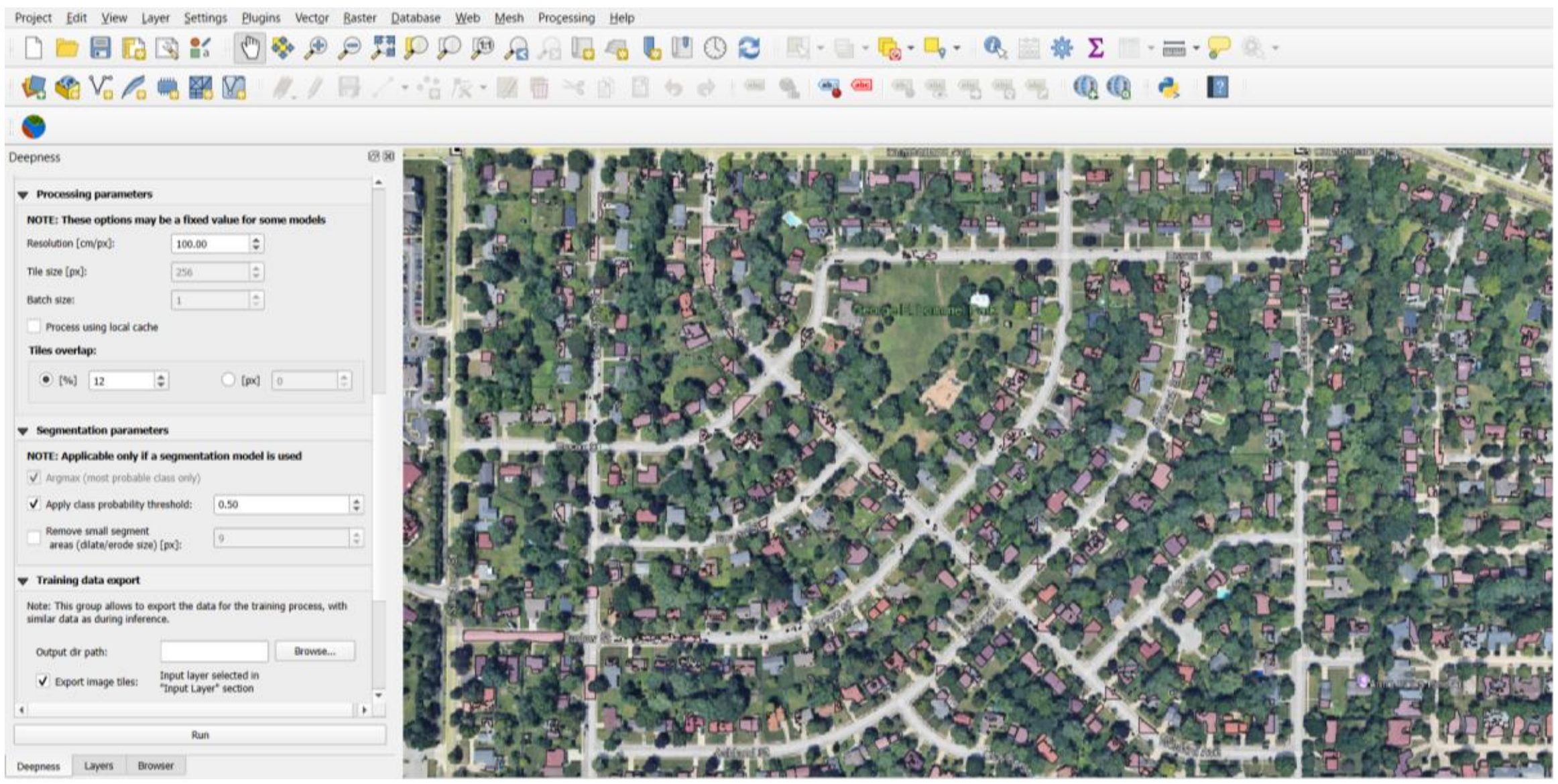

*Figure 1. AI-based digitization results with a 100 cm/px resolution*

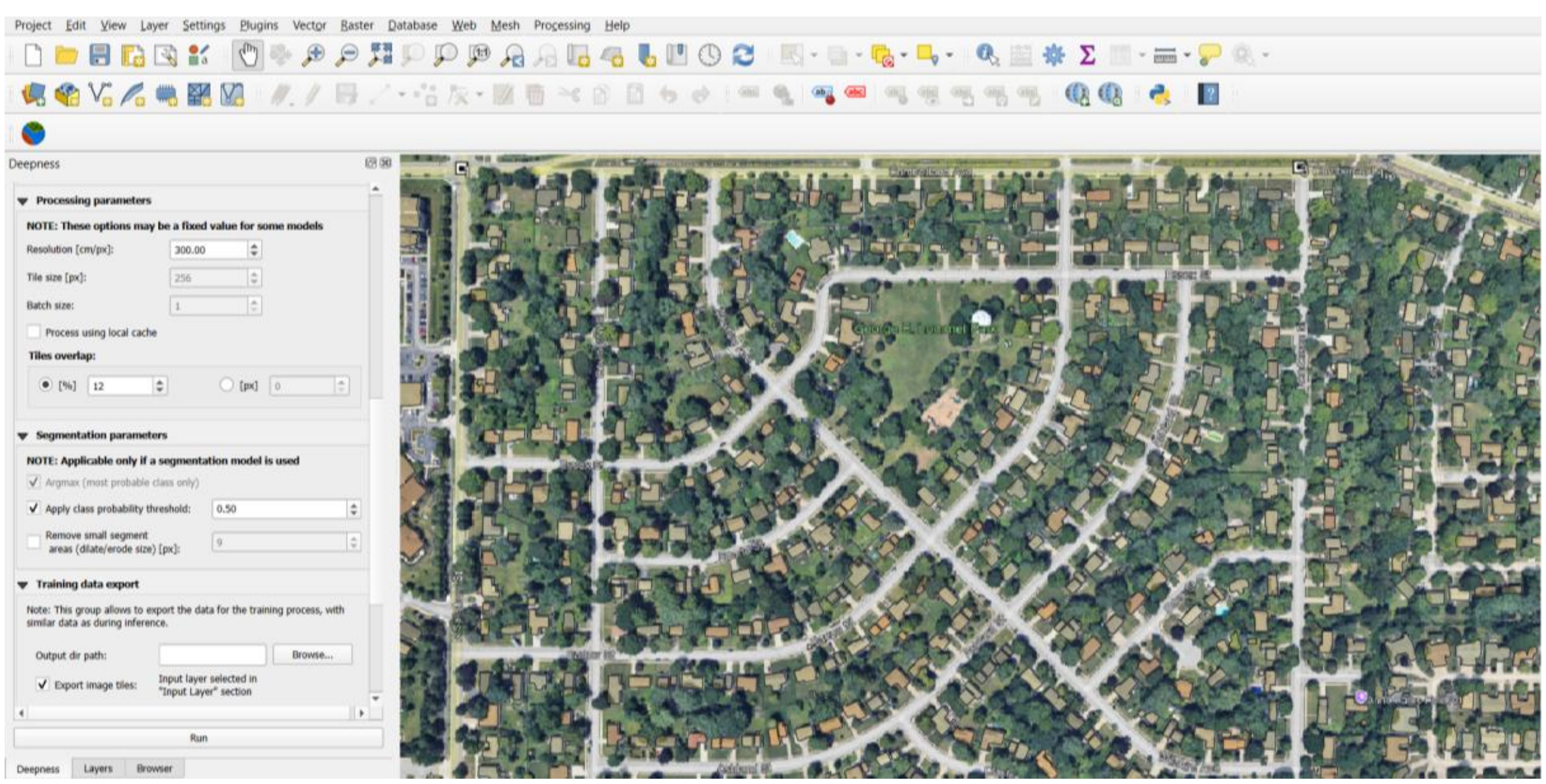

*Figure 2.  AI-based digitization results with a 300 cm/px resolution*

In addition, the percentage of tile overlap can significantly influence AI digitization performance (Cira et al, 2024). Using the same image and resolution, varying the tile overlap percentage leads to different building digitization results. At a resolution of 300 cm/px, setting the tile overlap to 0% yields 918 detected buildings as shown in Figure 3, whereas a 12% tile overlap results in 776 buildings as shown in Figure 4.

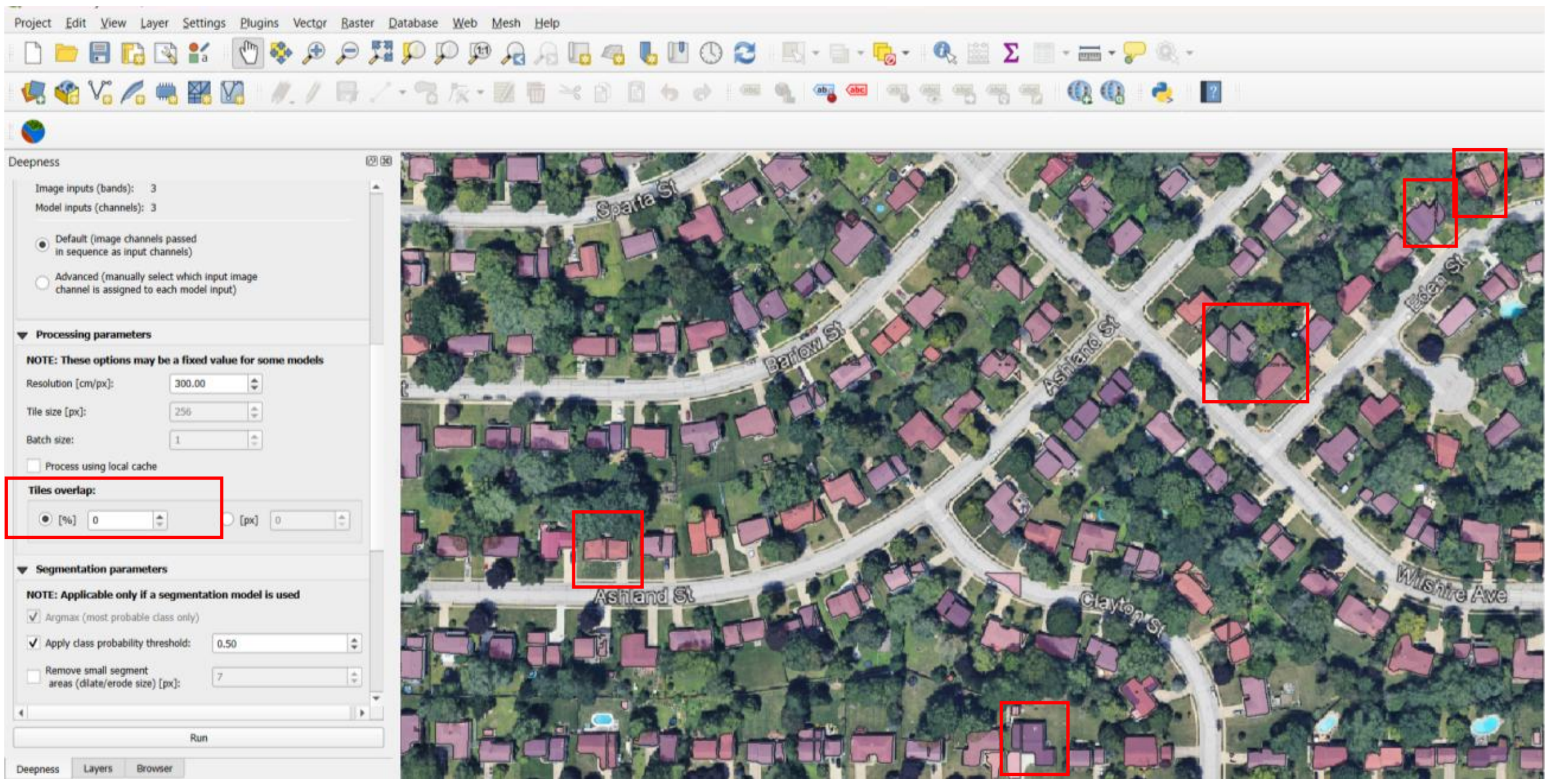

*Figure 3. AI digitization results in a 300 cm/px resolution & 0% of tiles overlap*

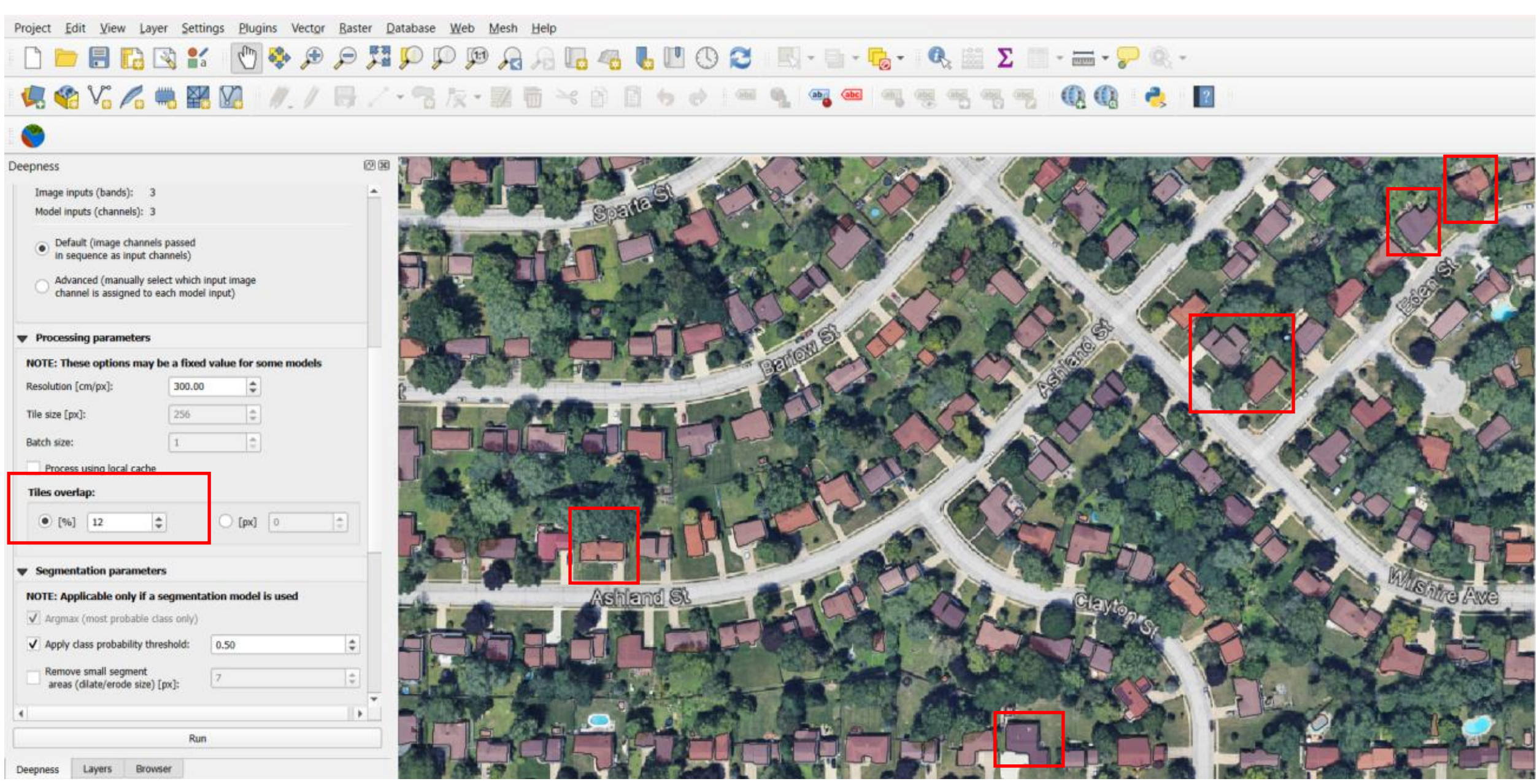

*Figure 4. AI digitization results in a 300 cm/px resolution & 12% of tiles overlap*

Table 1 shows that increasing the tile overlap percentage in the parameters can impact the number of buildings digitized. The variation in results is due to the impact of tile overlap on the segmentation process. A building that is split into two segments due to shadows or other factors may be merged. However, if the tile overlap percentage is too high, there is a risk of merging two adjacent buildings that should be considered separate (Cira et al, 2024). To identify the parameter that yields the best digitization results, a performance evaluation should be conducted.

*Table 1. Comparison of Buildings Digitized with Different Parameter Settings*

| Remove Small Segment | Resolution (cm/px) | Percentage Tiles Overlap | Number of Buildings Digitized |
|---|---|---|---|
| Disable | 300 cm/px | 0% | 918 |
| Disable | 300 cm/px | 12% | 776 |
| Enable | 300 cm/px | 12% | 611 |

In addition to the two parameters mentioned earlier, the segmentation parameter settings also influence the digitization results. When the remove small segment option is enabled, previously separated buildings due to small segments will merge into a single unit. In terms of the number of digitized buildings, enabling the removal of small segments with a resolution of 300 px/cm and a tile overlap percentage of 12% resulted in 611 buildings (Figure 5) being detected. In contrast, with the same resolution and overlap percentage but with remove small segment disabled, the number of detected buildings increased to 776 (Figure 6). This can be observed in several examples shown in the following figures.

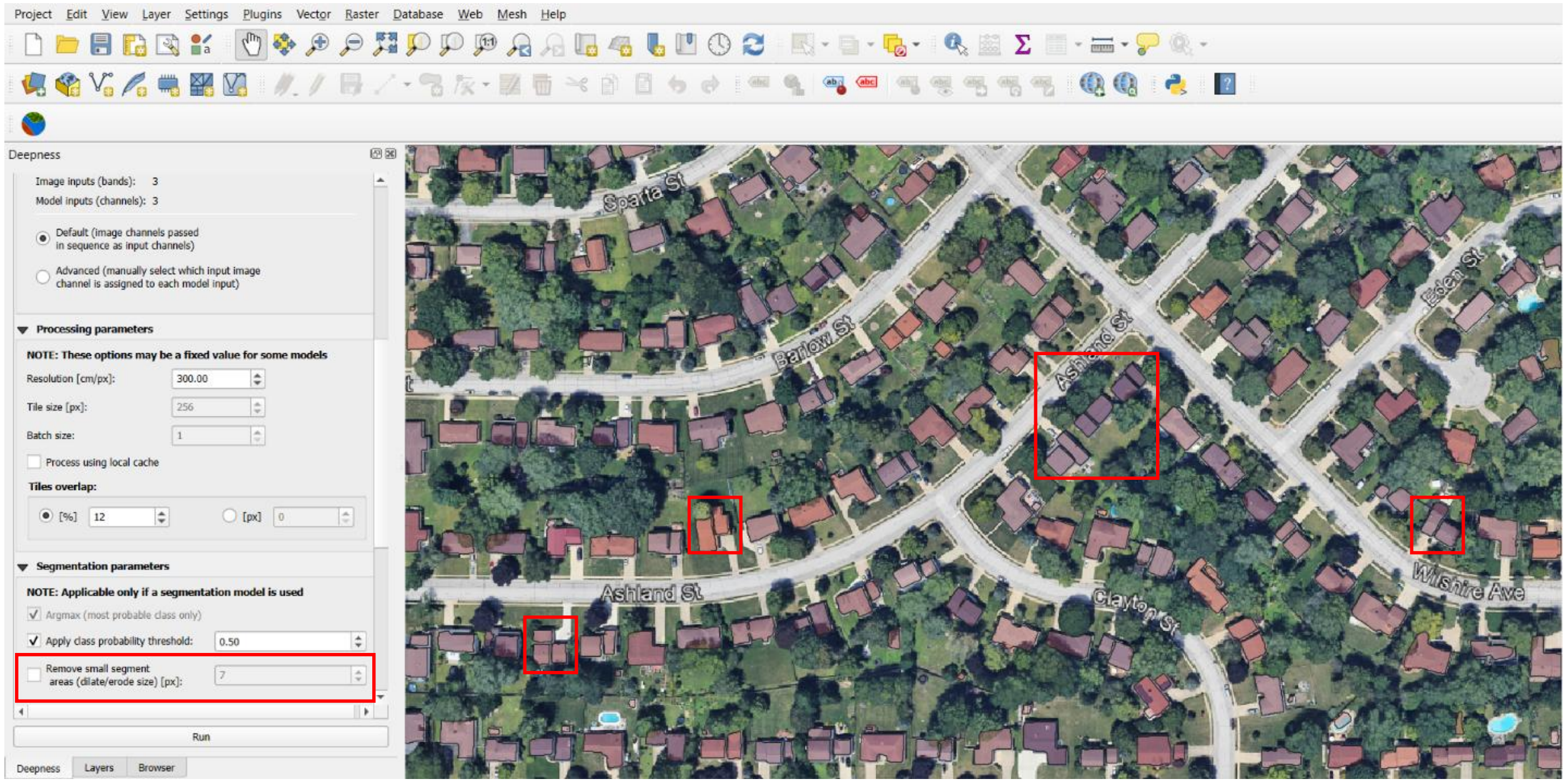

*Figure 5. AI digitization was performed at 300 px/cm resolution with 12% tile overlap, with remove small segment disabled.*

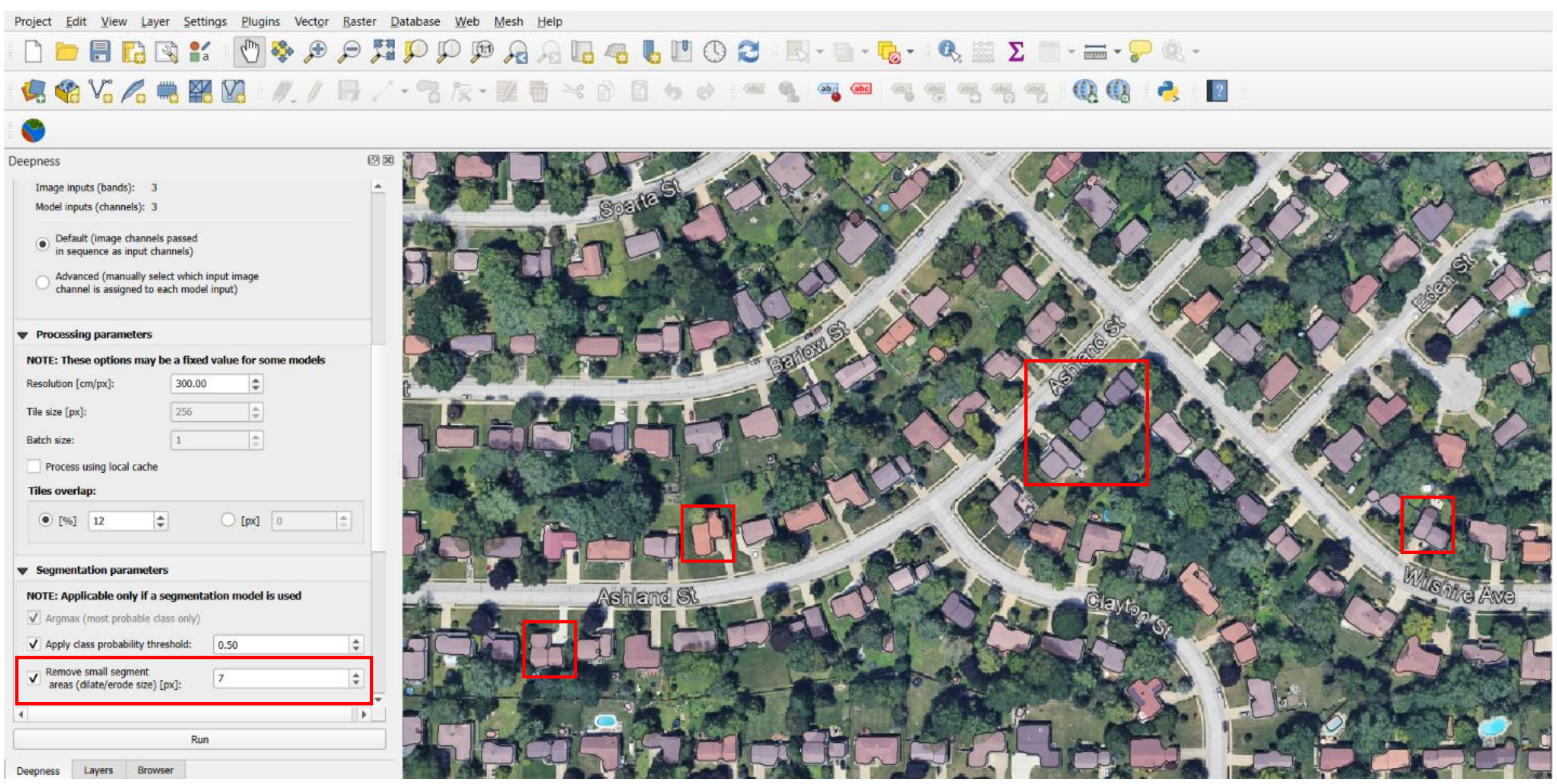

*Figure 6. AI digitization was performed at 300 px/cm resolution with 12% tile overlap, with remove small segment enabled.*

### *b. Effect of Parameter Settings on AI Building Segmentation Performance*

A study highlighted the effect of tile overlap on convolutional neural network performance for road classification, showing that a 12.5% overlap improved accuracy, while a 0% overlap resulted in lower performance (Cira et al., 2024). In this research, using the same resolution, the F1 scores were calculated with two different tile overlap percentages: 0% and 12%. The 12% tile overlap was selected, as the parameter does not support decimal values.

For this analysis, a subset of the digitized areas (buildings inside the red boundary shown in Figure 7) from both the AI and OSM datasets was selected for comparison. The analysis was performed using R software. Initially, the bounding box was set to define the analysis area, and the AI-generated digitization map was overlapped with the corresponding OSM map. This overlap allowed for the identification of areas where the two datasets intersected. The overlapping buildings serve as indicators of the similarity between the AI-based digitization results and the OSM map, providing valuable insights into the performance of the AI model in relation to the established OSM data.

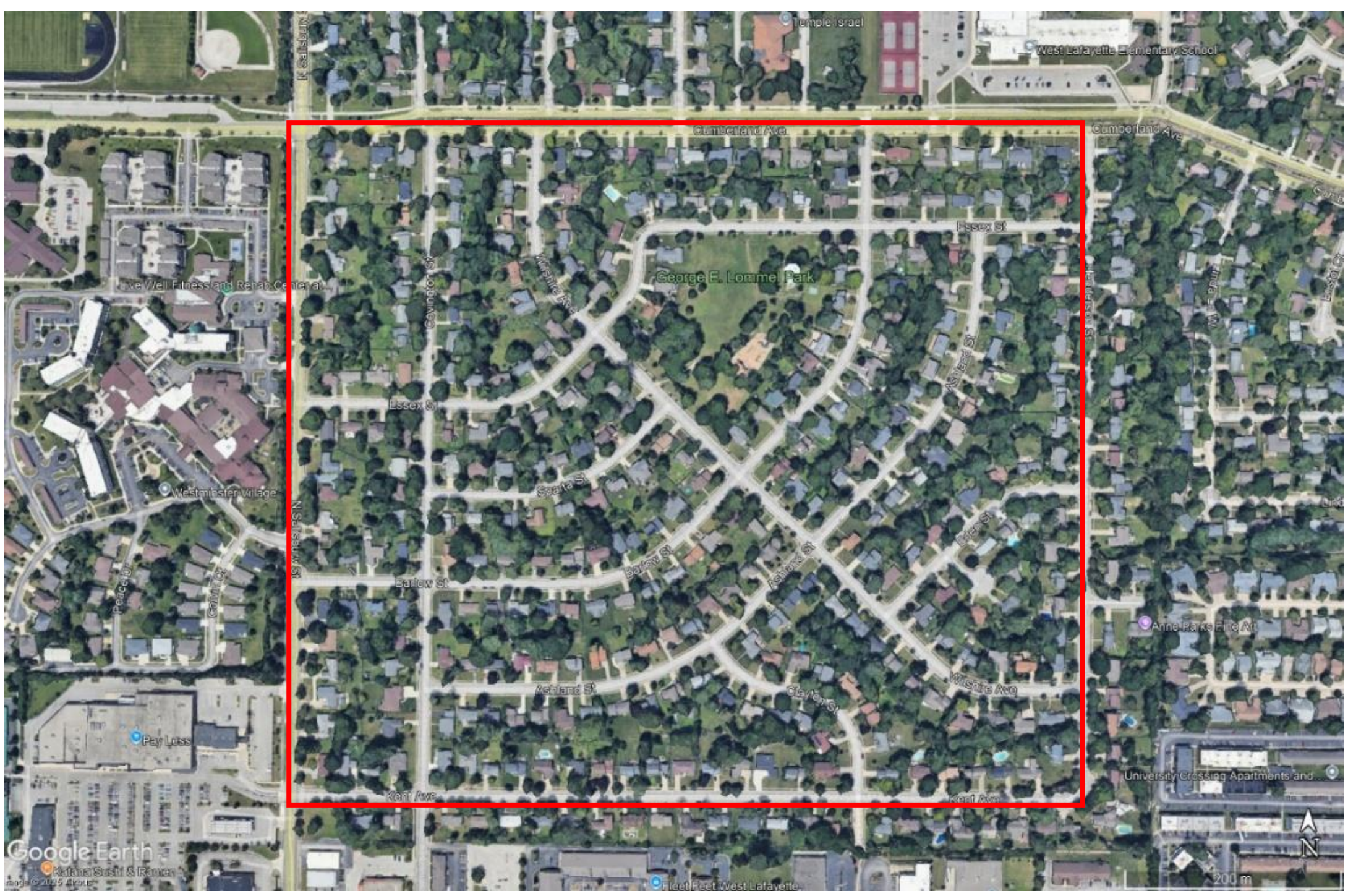

*Figure 7. Area of the research*

The performance of AI-based building detection in QGIS was compared with a dataset from OpenStreetMap (OSM), as presented in Table 2, Effect of Parameter Settings on AI Building Digitization. In the first test, OSM contained 320 buildings, and Deepness identified 603 buildings. A total of 222 buildings detected by the AI overlapped with the OSM buildings (IoU > 50%) and were classified as true positives. Deepness also detected buildings that did not match any OSM buildings and buildings with an IoU less than 50%. The number of false positives is 381 buildings, and the number of false negatives is 98 buildings.

In the second and third tests, the number of digitized and matched buildings changed significantly. In the second test, Deepness identified 497 buildings, with 257 matching the OSM ground truth. In the third test, it detected 398 buildings and had 282 matches out of the 320 buildings in the ground truth. The number of false positives decreased from 240 in the second test to 116 in the third test, indicating a notable reduction in irrelevant detections. The number of false negatives also decreased from 98 buildings in the first test to 38 buildings in the third test. Overall, the results demonstrate increasing performance in AI-based detection using QGIS, with the third test showing reductions in both false positives and false negatives.

*Table 2. Effect of Parameter Settings on AI Building Digitization*

| **Results** | **1st Test**<br>Resolution = 300 cm/px<br>Percentage Tiles Overlap = 0% | **2nd Test**<br>Resolution = 300 cm/px<br>Percentage Tiles Overlap = 12% | **3rd Test**<br>Resolution = 300 cm/px<br>Percentage Tiles Overlap = 12%<br>Remove Small Segment = Enable |
|---|---|---|---|
| Total number of buildings in OSM | 320 buildings | 320 buildings | 320 buildings |
| Total number of buildings identified by AI | 603 buildings | 497 buildings | 398 buildings |
| True Positive | 222 buildings | 257 buildings | 282 buildings |
| False Positive | 381 buildings | 240 buildings | 116 buildings |
| False Negative | 98 buildings | 63 buildings | 38 buildings |

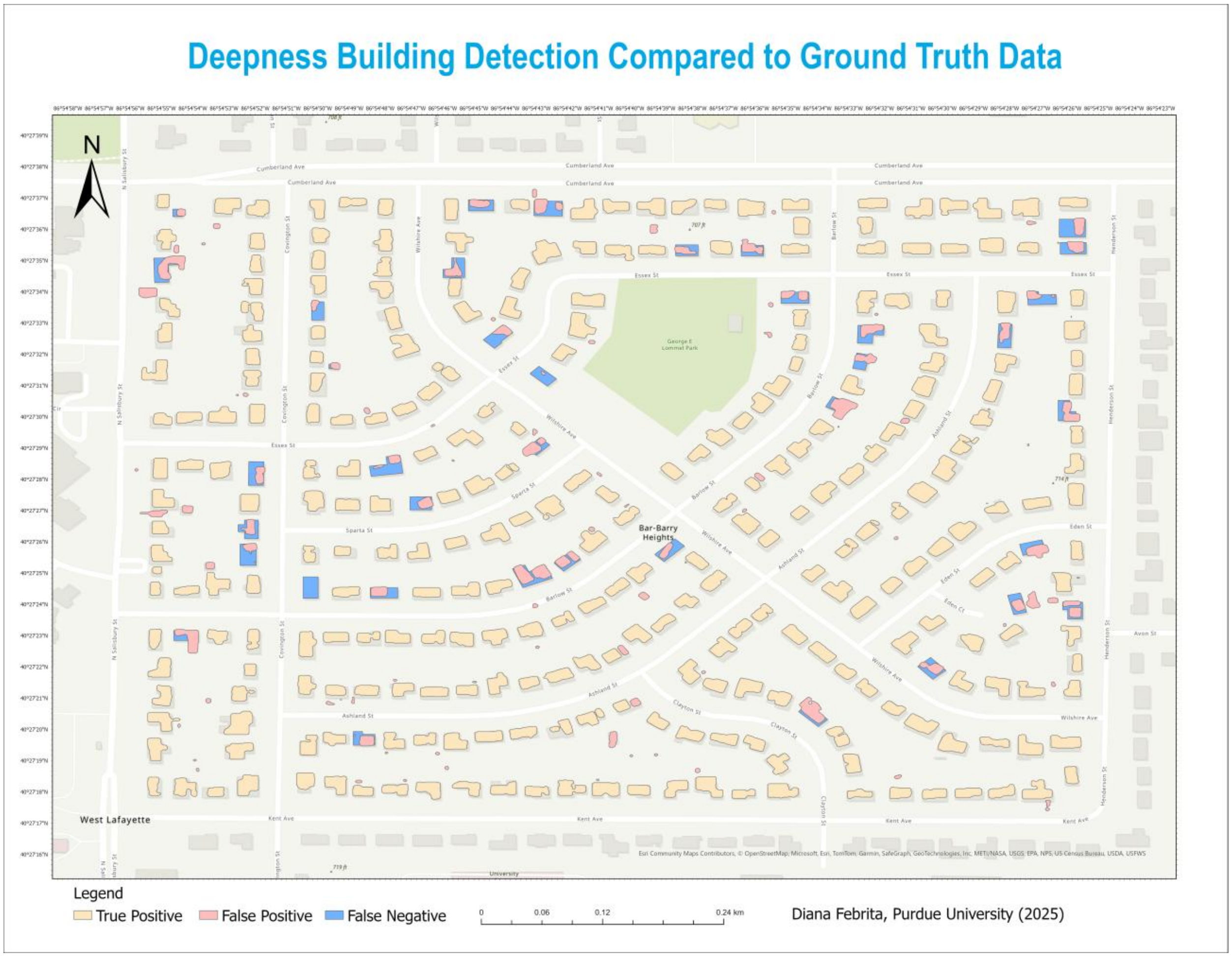

*Figure 8. Deepness Building Detection Compared to the Ground Truth Data*

Ghimire, Kim, and Acharya (2024) explain that the implementation of AI faces several challenges, such as model accuracy. This is consistent with the findings of this research, which shows that AI digitizes more buildings than humans do in OSM mapping. This happens because AI learns from a specific model and digitizes according to what it has learned. For example, AI may digitize storage buildings located behind the main buildings. Additionally, AI may also misclassify non-buildings as buildings or digitize a single building into two or more segments shown in Figure 9.

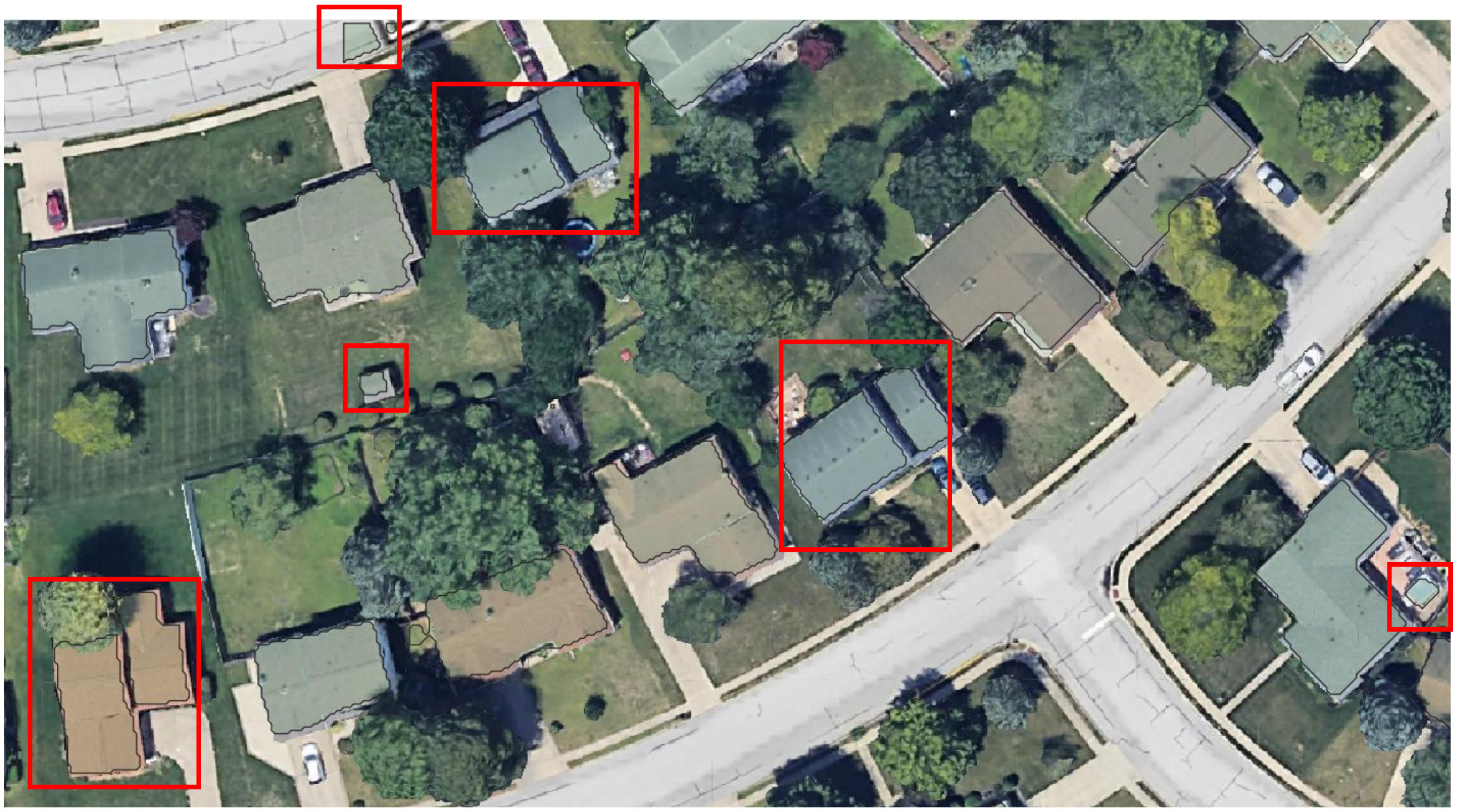

*Figure 9. AI Challenges in the Digitization Process*

To evaluate the AI's performance, the F1-score was calculated as shown in Table 3, Effect of Parameter Settings on AI-based Digitization Performance. The results indicate that the third test, using a 300 cm/px resolution, 12% tile overlap, and the "*Remove Small Segments*" option enabled, yielded the best overall performance. Precision increased from 36.82% in the first test and 51.71% in the second test to 70.85% in the third test. Meanwhile, recall reached 88.12% in the final test. These results demonstrate that parameter settings in Deepness have a substantial impact on AI performance.

Overall, the first test produced an F1-score below 50% (48.10%). However, adjustments to the resolution and tile overlap, along with enabling the segment removal setting, significantly improved performance. The F1-score increased to 62.91% in the second test and 78.55% in the third test. This improvement is supported by the reduction in both false positives and false negatives, which leads to higher precision and recall. The high recall observed in the second and third tests indicates that the model performs well in applications where missing objects must be minimized, such as detecting buildings in disaster response scenarios.

*Table 3. Effect of Parameter Settings on AI-based Digitization Performance*

| **Results** | **1st Test**<br>Resolution = 300 cm/px<br>Percentage Tiles Overlap = 0% | **2nd Test**<br>Resolution = 300 cm/px<br>Percentage Tiles Overlap = 12% | **3rd Test**<br>Resolution = 300 cm/px<br>Percentage Tiles Overlap = 12%<br>Remover Small Segment = Enable |
|---|---|---|---|
| Precision | 36.82 % | 51.71 % | 70.85 % |
| Recall | 69.38 % | 80.31 % | 88.12 % |
| F1-score | 48.10 % | 62.91 % | 78.55 % |

## Conclusion

In conclusion, the parameters set before digitizing with Deepness significantly influence the performance of the results. This study shows that as the resolution used for building segmentation changes, the number of detected buildings also changes. Additionally, this work reveals that the percentage of tile overlap affects both the number of detected buildings and the overall performance of the digitization process.

The model demonstrates strong overall performance, particularly in the third test. The high recall is especially beneficial in applications where missing targets could have serious consequences, such as disaster response. However, the precision results highlight areas that require further refinement to improve the model's accuracy.

## Acknowledgment

I would like to express my sincere gratitude to Professor Jie Shan for the invaluable feedback and insights in this research and the Geospatial Data Analytics class at Purdue University.